\documentclass{article}

\usepackage[preprint]{neurips_2026}

\usepackage{graphicx}
\usepackage{multirow}
\usepackage{xspace}
\usepackage[utf8]{inputenc} % allow utf-8 input
\usepackage[T1]{fontenc}    % use 8-bit T1 fonts
\usepackage{hyperref}       % hyperlinks
\usepackage{url}            % simple URL typesetting
\usepackage{booktabs}       % professional-quality tables
\usepackage{amsfonts}       % blackboard math symbols
\usepackage{nicefrac}       % compact symbols for 1/2, etc.
\usepackage{microtype}      % microtypography
\usepackage{xcolor}         % colors
 \usepackage{amsmath}

\usepackage{array}
\usepackage{caption}
\usepackage{makecell}
\title{ViCo3D: Empowering LiDAR-based Collaborative 3D Object Detection with Vision Foundation Models}

\author{%
  Haojie Ren, Songrui Luo, Lingfeng Wang, Yan Xia, \\
  Yao Li, Lu Zhang, Jiajun Deng, and Yanyong Zhang \\
  University of Science and Technology of China \\
  \texttt{rhj@mail.ustc.edu.cn}
}

\begin{document}
\newcommand{\systemname}{\textit{ViCo3D}\xspace}

\maketitle

\begin{abstract}

LiDAR-based collaborative 3D perception in Vehicle-to-Everything (V2X) systems typically relies on fusing bird's-eye-view (BEV) features across agents. However, current BEV representations, typically extracted by LiDAR backbones trained from scratch, are geometry-dominated and lack general semantic priors, inherently limiting the efficacy of feature-level collaboration.
Meanwhile, vision foundation models (VFMs) pretrained on large-scale image data have demonstrated strong capability in learning general-purpose and informative visual representations for 2D tasks, and have the potential to enhance agent-wise LiDAR BEV representations for collaboration.
Despite this potential, adapting VFMs to LiDAR-based 3D detection remains challenging due to the substantial image–point cloud modality gap. 
To bridge this gap, we propose ViCo3D, a collaborative 3D object detection framework powered by VFMs. Specifically, ViCo3D adapts VFMs to LiDAR-based collaborative perception from three aspects:
First, ViCo3D projects point clouds onto the BEV plane as three-channel images, enabling DINOv2 to extract BEV-space visual features from LiDAR inputs.
Besides, to effectively integrate these DINOv2-derived features with LiDAR geometric features, ViCo3D introduces a multi-scale BEV fusion module within the single-agent encoder.
In addition, ViCo3D adopts an ego-centric cross-agent fusion strategy to aggregate complementary information from multiple agents.
Experiments on DAIR-V2X and V2XSet demonstrate that ViCo3D achieves state-of-the-art 3D detection performance. Remarkably, it delivers up to $1.8\times$ greater collaborative gains than prior methods on DAIR-V2X. The code will be made public available for future investigation.

\end{abstract}

\section{Introduction}

% 车路协同感知系统通过在路侧部署感知设备，并结合车端与路侧之间的信息交互，实现对交通场景的联合建模与环境感知。相较于传统单车感知模式，该范式能够利用车端与路侧在观测视角和感知范围上的互补性，为车辆提供更加稳定、全面的环境信息，从而有效缓解单车感知在遮挡、远距离目标探测及复杂交通场景理解等方面的局限。因此，车路协同感知被认为是实现高等级自动驾驶的重要技术方向之一。
% 尽管如此，车路协同感知相较于单车3D感知仍面临更大的技术挑战。特别是由于车端与路侧在传感器配置、观测视角及数据分布等方面存在显著差异，二者提取的特征往往具有较强的异质性，这在一定程度上制约了协同融合模块的性能上限。然而，现有研究大多关注协同框架设计与融合策略优化，对跨主体特征分布差异及其对融合效果的影响关注不足。基于此，本文重点研究如何学习对协同融合更加友好的车端与路侧特征表示，从而提升车路协同感知系统的整体感知性能。

\begin{figure*}[t]
\centering
\begin{minipage}[t]{0.62\textwidth}
    \vspace{0pt}
    \centering
    \includegraphics[width=\linewidth]{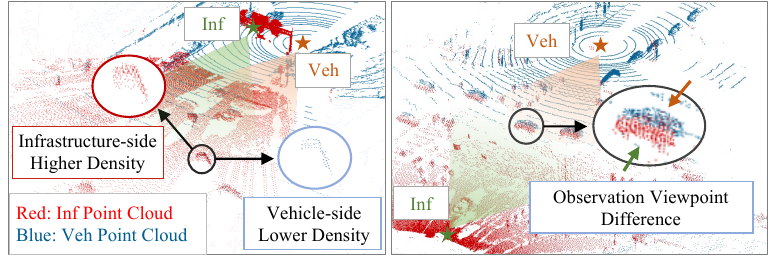}
    \captionof{figure}{Density Distribution and Viewpoint Discrepancies Between Veh-Inf Point Clouds.}
    \label{fig:difference_of_vi}
\end{minipage}
\hfill
\begin{minipage}[t]{0.35\textwidth}
    \vspace{1pt}
    \centering
    \captionof{table}{Our method yields lower cross-view feature similarity and higher detection performance.}
    \label{tab:cos_sim}
    % \vspace{-0.3em}
    
% \centering
% % \caption{Cross-view feature similarity comparison using cosine similarity. }
% \label{tab:region_similarity_cos}
% \begin{tabular}{lcc}
% \toprule
% Region & Baseline & Ours \\
% \midrule
% All       & \textbf{0.8243} & 0.6622  \\
% FG-Both   & \textbf{0.8316} & 0.5140  \\
% FG-Union  & \textbf{0.7537} & 0.4858  \\
% AP@0.3    &  82.65  &   \textbf{86.35} \\
% AP@0.7    &  64.73  &   \textbf{71.39} \\
% \bottomrule
% \end{tabular}

\centering
% \caption{Our method yields lower cross-view feature similarity and higher detection AP.}
\label{tab:cos_sim_ap}
\begin{tabular}{lcc}
\toprule
Metric & Baseline & Ours \\
\midrule
% \multicolumn{3}{c}{Cross-view feature similarity $\downarrow$} \\
FG-Both  & 0.8316 & 0.5140 \\
FG-Union & 0.7537 & 0.4858 \\
\midrule
% \multicolumn{3}{c}{Detection performance $\uparrow$} \\
AP@0.3 $\uparrow$     & 0.8265 & \textbf{0.8635} \\
AP@0.7 $\uparrow$     & 0.6473 & \textbf{0.7139} \\
\bottomrule
\end{tabular}
\end{minipage}
\vspace{-0.5cm}
\end{figure*}

LiDAR-based V2X collaborative perception methods typically rely on fusing bird’s-eye-view (BEV) features extracted by LiDAR backbones from the point clouds of multiple agents. These backbones are commonly trained from scratch on task-specific 3D detection datasets and mainly encode local point-cloud geometry. However, different agents often exhibit substantial discrepancies in sensor configurations, observation perspectives, and data distributions, as shown in Fig.~\ref{fig:difference_of_vi}. 
Such discrepancies can lead to heterogeneous BEV feature representations, thereby constraining the effectiveness of feature-level collaborative fusion.
% Vehicle-to-Everything (V2X) collaborative 3D detection based on LiDAR enables multiple traffic agents to jointly perceive surrounding objects by sharing complementary sensory information. 
% Compared with single-vehicle perception, it alleviates perception challenges caused by occlusion and limited detection range, making it a promising paradigm for high-level autonomous driving. 
% However, different agents often exhibit substantial discrepancies in sensor configurations, observation perspectives, and data distributions, as shown in Fig.~\ref{fig:difference_of_vi}. 
% Such discrepancies can lead to heterogeneous BEV feature representations, thereby constraining the effectiveness of collaborative fusion.
% However, different agents often exhibit substantial discrepancies in sensor configurations, observation perspectives, and data distributions, as shown in Fig.~\ref{fig:difference_of_vi}. 
% Such cross-agent discrepancies can lead to heterogeneous feature representations, thereby constraining the performance of feature-level collaborative fusion.
Existing studies have not sufficiently addressed this issue. 
This paper focuses on mitigating cross-agent feature heterogeneity by learning more fusion-friendly BEV representations, thereby improving collaborative 3D detection performance.

% Therefore, this paper aims to learn a shared feature space that better facilitates the fusion of vehicle-side and infrastructure-side features, thereby improving the final perception performance.

% \begin{figure*}
%     \centering
%     \includegraphics[width=0.8\linewidth]{image/first_0410.pdf}
%     \caption{Density Distribution and Viewpoint Discrepancies Between Vehicle-Side and Infrastructure-side Point Clouds}
%     \label{fig:difference_of_vi}
%     \vspace{-0.3cm}
% \end{figure*}

% 针对车端与路侧之间的异构性，现有方法大致可分为两类。第一类采用后融合策略，车辆与路侧单元主要交换目标框、类别等高层语义信息。然而，这类方法高度依赖各自的感知结果，一旦目标未被单侧正确感知，后续融合过程通常难以恢复相应信息，因此其协同增益上限受到限制。另一类方法以 DI-V2X 为代表，试图通过特征蒸馏缩小车端与路侧之间的表征差异。然而，这类方法更强调特征一致性，可能在一定程度上忽略了车路协同中由不同视角带来的互补信息，从而限制了融合性能的进一步提升。
Existing feature-level V2X collaborative perception methods mainly focus on learning consistent BEV features across agents to improve collaborative perception performance.
For example, DI-V2X (~\cite{li2024di}) uses feature distillation to encourage different agents, such as vehicles and infrastructure, to learn more consistent feature representations. 
However, we argue that collaborative perception should not only pursue feature consistency, but also preserve detection-relevant complementary information introduced by diverse observations. 
Tab.~\ref{tab:cos_sim} provides empirical support for this view: \systemname produces lower cross-view feature similarity than the baseline, yet achieves better detection performance. 
% This suggests that higher feature similarity is not always necessary for better collaborative detection, highlighting the importance of considering complementary information in feature-level fusion.
However, existing LiDAR backbones may be insufficient for learning such detection-relevant complementary representations, as they are trained from scratch on task-specific 3D detection datasets with limited scale and diversity.
Although point-cloud foundation models offer a direct route, they remain limited by the sparse, irregular, and sensor-dependent nature of LiDAR data.

As an alternative, we observe that BEV projection provides an image-like interface: projecting point clouds onto the BEV plane produces structured 2D maps that preserve spatial layouts and geometric patterns. 
Meanwhile, vision foundation models (VFMs) have demonstrated strong capability in extracting rich and diverse visual features through large-scale image pretraining. This motivates us to explore whether VFMs can be adapted to enhance agent-wise LiDAR BEV representations for collaboration.
However, as shown in Tab.~\ref{tab:diff_branch}, using DINOv2-derived BEV features alone leads to severe performance degradation. This indicates that directly applying VFMs to LiDAR-based 3D detection is non-trivial due to the image--point cloud modality gap.
Therefore, instead of replacing LiDAR backbones with VFMs, we design a fusion framework that integrates VFM-derived features with LiDAR BEV features.
Nevertheless, this fusion remains challenging. Attention-based fusion is a natural choice for modeling interactions between different feature sources, but directly applying dense attention on BEV feature maps incurs prohibitive memory and computational costs due to the large number of spatial tokens. 

To address these challenges, we propose \systemname, a VFM-enhanced framework for LiDAR-based collaborative 3D detection. \systemname projects point clouds onto the BEV plane as three-channel images, enabling DINOv2 to extract BEV-space visual features from LiDAR inputs. To efficiently integrate these DINOv2-derived features with LiDAR geometric features, \systemname introduces a multi-scale BEV fusion module within the single-agent encoder. It first applies a Low-Resolution Global Fusion (\textbf{LRG}) module to capture global contextual interactions, and then uses a High-Resolution Refinement Fusion (\textbf{HRR}) module to refine local spatial details at a finer BEV scale.
Finally, after receiving data from other agents, the ego vehicle fuses them under an ego-centric paradigm.

Overall, \systemname is designed to better exploit complementary information across agents by integrating DINOv2-based visual representations with point-cloud-based BEV features. In this way, \systemname provides richer and more informative representations for vehicle-infrastructure collaborative perception.
We conduct extensive experiments on the real-world DAIR-V2X (~\cite{li2024di}) dataset and the simulated V2XSet (~\cite{xu2022v2x}) dataset. Experimental results show that \systemname achieves the best performance among BEV-based detection methods, reaching 82.80/71.39 BEV AP@0.5/0.7 on DAIR-V2X and 97.20/93.23 on V2XSet. Moreover, \systemname achieves the largest collaborative gain. At IoU = 0.5 on DAIR-V2X, it improves the single-vehicle baseline by 13.01 percentage points, which is about 1.89$\times$ that of INSTINCT (~\cite{xu2025instinct}) and 1.60$\times$ that of DI-V2X (~\cite{li2024di}).

% 我们的贡献如下所示：
% Our contributions can be summarized as follows:
% \begin{itemize}
%     \item To the best of our knowledge, this is the first work to introduce vision foundation models into vehicle-infrastructure collaborative perception to enhance feature representations. We demonstrate that vision foundation models can enrich semantic information and produce representations that are more suitable for collaborative fusion.
    
%     \item We present a two-stage feature optimization framework for vehicle-infrastructure collaborative perception. In the local perception stage, a semantic branch is introduced to complement the original PointPillars geometric branch, along with a Low-Resolution Global Fusion (\textbf{LRG}) module and a High-Resolution Refinement Fusion (\textbf{HRR}) module for effective multi-scale feature enhancement. In the collaborative fusion stage, we further propose an Ego-Centric Collaborative Feature Fusion (\textbf{ECC}) mechanism to selectively integrate complementary information from other agents while preserving the local perception advantages of the ego vehicle.
    
%     \item Extensive experiments on the DAIR-V2X dataset show that our method achieves state-of-the-art collaborative perception performance and delivers the largest collaborative gain over the single-vehicle baseline.
% \end{itemize}

\section{Related Work}

\subsection{Collaborative Perception}

Collaborative perception enables multiple agents to share complementary information while alleviating occlusion and limited sensing range. 
% 依据agent之间数据交换类型的不同，V2X 协同感知方法大致可以分为三类：早期融合、中间融合和后期融合。早期融合模型在原始数据层面整合来自不同智能体的信息，以获取整体感知。然而，在智能体之间传输原始 3D 数据不可避免地会消耗更多通信带宽，这阻碍了其在真实场景中的应用。相比之下，后期融合方法仅在感知结果层面进行协同，例如融合各智能体预测得到的 3D 边界框，因此能够显著降低通信负担。但这类方法高度依赖单个智能体的检测结果，当局部感知存在噪声、漏检或智能体间位姿误差时，最终融合性能容易受到影响。中间融合的方法在特征层面进行融合，它们只在智能体之间传输包含关键信息的压缩特征图，能够平衡传输信息量以及通信带宽，且其对定位误差更加鲁班，因而受到了广泛的关注。
Depending on the type of data exchanged among agents, V2X collaborative perception methods can be broadly categorized into three groups: early fusion, intermediate fusion, and late fusion. 
Early fusion aggregates 3D point clouds from different agents to obtain a more complete scene representation (\cite{arnold2020cooperative}\cite{chen2019cooper}\cite{chen2019f}\cite{yu2022dair}).  However, raw point cloud data are large in volume, and transmitting them incurs substantial communication overhead, which limits their application in real-world V2X scenarios.
In contrast, late fusion approaches (\cite{li2022v2x}\cite{xu2022opv2v}\cite{yu2023v2x}) perform collaboration only at the result level, such as the 3D bounding boxes predicted by individual agents, thereby significantly reducing communication costs. However, their performance heavily depends on the perception quality of individual agents and is also sensitive to pose errors between agents.
Intermediate fusion methods (\cite{liu2020when2com, liu2020who2com, xu2022v2x, chen2019cooper, hu2024communication}) provide a balanced strategy between the amount of exchanged information and communication bandwidth. By transmitting only compressed feature maps that contain task-critical information across agents, these methods significantly reduce communication costs while remaining more robust to pose errors than the other two types.

In this paper, we focus primarily on intermediate-fusion approaches. V2X-ViT~(\cite{xu2022v2x}) employs heterogeneous multi-agent self-attention to adaptively aggregate BEV features from different agents. Where2comm~(\cite{hu2022where2comm}) reduces bandwidth by transmitting only spatially critical regions guided by confidence maps. CodeFilling~(\cite{hu2024communication})extends this idea by introducing a codebook-based representation. V2X-PC~(\cite{liu2024v2x}) departs from dense BEV feature transmission by communicating compact point clusters. 

Besides extensive efforts on communication optimization, a few studies have also investigated the data distribution discrepancies across agents, especially between vehicle-side and infrastructure-side agents. DI-V2X~(\cite{li2024di}) adopts a distillation-based framework to align feature representations across agents. While this strategy helps reduce the representation discrepancy among heterogeneous agents, how to sufficiently preserve and exploit the complementary information in their observations remains underexplored. To this end, we propose \systemname, which not only mitigates distribution discrepancies among agents but also fully exploits their complementary information.

\subsection{Vision and 3D Foundation Models for 3D Data}

Most existing 3D foundation models (\cite{wu2024point}\cite{wu2025sonata}) are trained on small-scale point cloud datasets and primarily target object-level tasks, which limits their generalization to large-scale outdoor 3D perception.
On the contrary, recent vision foundation models (VFMs) (\cite{kirillov2023segment}\cite{radford2021learning}), such as DINOv2 (\cite{oquab2023dinov2}) and MAE (\cite{he2022masked}), demonstrate strong transferability due to large-scale self-supervised pretraining on massive image datasets. Driven by the representational strength of VFMs, recent studies have investigated transferring visual pretraining to point cloud understanding. 

Existing studies mainly explore two paradigms for transferring VFMs to point cloud understanding. 
The first is feature distillation, where pretrained visual models act as teachers to guide 3D representation learning. 
Representative methods include Seal (\cite{liu2023segment}), which distills knowledge from pretrained visual models into point cloud sequences, and HVDistill (\cite{zhang2024hvdistill}), which aligns image and point cloud representations across multiple 2D projections through unsupervised hybrid-view distillation. 
The second paradigm projects point clouds into image-like representations, such as bird’s-eye view (BEV) (\cite{luo2023bevplace}) or range image view (RIV) (\cite{chen2021overlapnet}), allowing VFMs to be directly reused for LiDAR-based perception tasks. 
For example, ImLPR (\cite{jung2025imlpr}) projects LiDAR data into RIVs and applies vision-pretrained backbones for LiDAR place recognition (LPR) without relying on RGB images.

In this work, we are the first to explore the potential of vision foundation models for enhancing collaborative perception.

\section{Method}

\subsection{Overall Framework}

\begin{figure*}[h]
  \centering
  \includegraphics[width=0.95\linewidth]{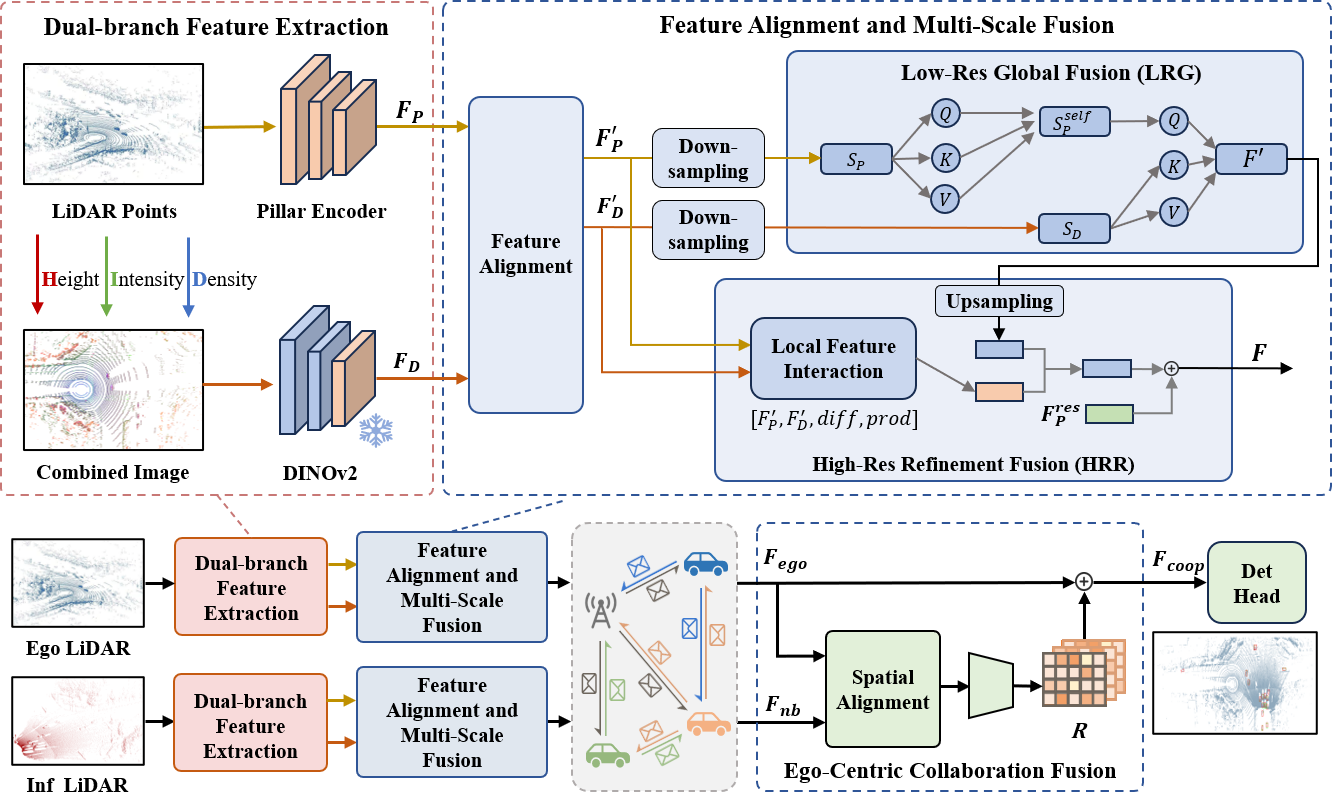}
  \caption{The Framework of ViCo3D.}
  \vspace{-0.5cm}
  \label{fig:framework}
\end{figure*}

% 在车路协同感知任务中，车端和路端智智能体能够从不同视角对场景进行互补的观测，然而由于传感器位姿、观测视角以及遮挡模式的差异，它们通常表现出不同的点云分布。因此，如何克服智能体之间的表示差异，并保留他们之间的互补线索，是V2X感知中的研究难题。
In V2X collaborative perception, different agents can provide complementary observations of the scene from diverse perspectives. However, due to differences in sensor poses, viewpoints, and occlusion patterns, they usually exhibit distinct point cloud distributions. How to overcome the representation discrepancies among agents while preserving their complementary cues remains a challenging problem in V2X perception.

To address this problem, we propose \systemname, a DINOv2-guided framework for V2X collaborative perception. 
The core idea is to leverage the transferable representation capability of pretrained DINOv2 to enhance the BEV feature space of heterogeneous agents. 
As shown in Fig.~\ref{fig:framework}, \systemname contains three modules.
First, the dual-branch feature extraction module (Sec.~\ref{sec:representation_con}) extracts two types of BEV features for each agent. 
Second, the feature alignment and multi-scale fusion module (Sec.~\ref{sec:feature_fusion}) integrates DINOv2 features with LiDAR BEV features. In this module, we first align the features encoded by the two branches to reduce their representation mismatch. 
We then propose a multi-scale fusion paradigm to integrate global context and local spatial details, with low-resolution fusion for global feature interaction (LRG) and high-resolution fusion for local feature refinement (HRR).
Finally, in the ego-centric collaboration fusion module (Sec.~\ref{sec:ma_fusion}), the enhanced BEV features from collaborating agents are transformed into the ego coordinate system and aggregated for final 3D object detection.

\subsection{Dual-branch Feature Extraction} 
\label{sec:representation_con}

In this paper, we leverage a pretrained vision model, DINOv2, to enhance LiDAR-based 3D perception. However, we still face two key challenges: 1) DINOv2 is designed for image inputs and cannot directly process irregular LiDAR point clouds; 2) as shown in Tab.~\ref{tab:diff_branch}, directly using DINOv2 features leads to suboptimal performance, because they are not specifically optimized for localization-sensitive 3D object detection. 
To address these challenges, we adopt a dual-branch encoding framework. 

\subsubsection{Pillar Branch}

In the pillar branch, we adopt a PointPillars-based encoder~(\cite{lang2019pointpillars}) to extract LiDAR-based BEV features. Specifically, each agent first voxelizes its point cloud into vertical pillars in the BEV space. Each pillar is encoded into a feature vector, and the resulting features are scattered onto the BEV grid according to their spatial locations. The output is a dense LiDAR BEV feature map \(F_{P} \in \mathbb{R}^{C \times H \times W}\), where \(C\) denotes the number of feature channels, and \(H\) and \(W\) denote the height and width of the BEV feature map, respectively.

\subsubsection{DINOv2 Branch}

As DINOv2 is designed for image inputs and cannot directly process irregular point clouds, we convert each raw LiDAR point cloud into a  BEV image.
Specifically, given a point cloud \(P=\{p_i\}_{i=1}^{N}\), each point is represented as \(p_i=(x_i, y_i, z_i, I_i)\), where \(x_i, y_i, z_i\) denote the 3D coordinates of the \(i\)-th point and \(I_i\) denotes its intensity. We first filter out points outside the predefined 3D region of interest, defined by \([x_{\min}, x_{\max}]\), \([y_{\min}, y_{\max}]\), and \([z_{\min}, z_{\max}]\). The remaining points are then projected onto a regular BEV grid with spatial resolution \(r\). For each point \(p_i\), its corresponding BEV pixel coordinates are computed as \(u_i = \left\lfloor (x_i - x_{\min}) / r \right\rfloor\) and \(v_i = \left\lfloor (y_{\max} - y_i) / r \right\rfloor\). 

After mapping all valid points onto the BEV grid, we construct a three-channel BEV image. 
The first channel records the maximum normalized height of points within each grid cell, capturing local vertical geometric structures.
The second channel stores the normalized reflectance intensity, providing complementary surface property information.
The third channel represents the point density of each cell, where the raw point count is further normalized using logarithmic scaling to alleviate the imbalance caused by non-uniform point distributions. 
Finally, the three channels are stacked into a structured BEV image and fed into the DINOv2 branch, producing a BEV feature map \(F_{D} \in \mathbb{R}^{D \times H \times W}\), where \(D\) denotes the number of feature channels.

\subsection{Feature Alignment and Multi-Scale Fusion} \label{sec:feature_fusion}

After obtaining the PointPillars branch feature \(F_P\) and the DINOv2 branch feature \(F_D\), we introduce a feature alignment and multi-scale fusion module to construct a unified and robust fused representation.

\subsubsection{Feature Alignment}
Since \(F_P\) and \(F_D\) are derived from different input forms and learning objectives, their feature spaces may differ. 
Specifically, the PointPillars branch encodes raw point clouds for 3D detection, whereas the DINOv2 branch extracts features from BEV images based on visual pretraining. Therefore, direct fusion without alignment may make it difficult to effectively exploit their complementary information.
To mitigate this issue, we first use two learnable projection layers to map \(F_P\) and \(F_D\) into a shared feature space, obtaining the aligned features \(F'_P=\phi_P(F_P)\) and \(F'_D=\phi_D(F_D)\), respectively. Here, \(\phi_P(\cdot)\) and \(\phi_D(\cdot)\) denote learnable projection functions implemented by a \(1 \times 1\) convolution followed by normalization and a nonlinear activation. 

\subsubsection{Low-Res Global Fusion (LRG)}

Considering that BEV feature maps are dense, directly applying transformer-based interaction at the original resolution would incur substantial computational and memory overhead. As shown in Tab.~\ref{tab:downsample}, when the feature map size is \(112 \times 252\), each branch contains \(28{,}224\) tokens, leading to an attention score matrix that requires up to \(48.62\) GB of memory in FP32. In contrast, when the downsampling ratio is \(4\times\), the memory cost is reduced to \(1/256\) of that at the original resolution, i.e., approximately 0.186 GB, making the overhead acceptable. Therefore, we perform global feature interaction on \(4\times\)-downsampled features.

Specifically, the low-resolution global fusion proceeds as follows.
We first downsample the aligned features \(F'_P\) and \(F'_D\), and flatten them into token sequences \(S_P\) and \(S_D\), respectively.
To preserve the BEV spatial layout after flattening, we add 2D positional embeddings to the tokens. 
We then apply self-attention on \(S_P\), producing \(S_P^{{self}}\), to model long-range spatial dependencies within the detection-oriented BEV representation.
After that, cross-attention is performed in a PointPillars-dominant manner, where \(S_P^{{self}}\) serves as the query and \(S_D\) serves as the key and value, producing the fused token sequence \(S_P^{{fuse}}\).
This design preserves the task-specific structure of the PointPillars feature stream while selectively aggregating complementary information from DINOv2 features.
Finally, \(S_P^{{fuse}}\) is reshaped into a 2D feature map and upsampled to the original BEV resolution, yielding the low-resolution global fusion feature \( F^{'} \).

\subsubsection{High-Res Refinement Fusion (HRR)}

However, as shown in Tab.~\ref{tab:ablation_main}, performing feature fusion only at a low resolution facilitates global information integration and significantly improves recall, but leads to a 5.4\% drop in detection precision. We attribute this to the fact that low-resolution interaction inevitably discards part of the fine-grained local details, especially those in the DINOv2 features, thereby weakening the representation of object boundaries and local structures. 
Motivated by this observation, we further introduce a high-resolution local refinement stage after low-resolution global fusion to compensate for the loss of local information.

In this stage, we directly model the aligned features \(F'_P\) and \(F'_D\) at the original BEV resolution. 
Although attention-based interaction is effective for feature fusion, applying it to high-resolution dense BEV features would incur prohibitive memory overhead. 
Therefore, we explicitly construct lightweight local feature relationships between the PointPillars and DINOv2 features. 
Specifically, beyond simple concatenation, we introduce two interaction terms, \(|F'_P - F'_D|\) and \(F'_P \odot F'_D\). 
The former captures local discrepancies between the two feature sources, while the latter highlights their complementary activation regions. 
These features are then concatenated and processed by a lightweight convolutional module to produce a high-resolution refinement feature \( F_{local} \).

Finally, we concatenate \(F'\) and \(F^{{local}}\), and feed them into a lightweight convolutional layer to generate the multi-scale fused feature \(F^{{ms}}\). 
In addition, to maintain the stability of the detection-oriented feature stream, we further introduce a residual pathway from \(F'_P\). 
A spatial dynamic gate is then used to adaptively balance \(F^{{ms}}\) and the PointPillars residual:
\begin{equation}
F = \sigma\left(\psi\left([F^{ms}, F'_P]\right)\right) \odot F^{ms}
+ \left(1-\sigma\left(\psi\left([F^{ms}, F'_P]\right)\right)\right) \odot F'_P ,
\end{equation}
where $[\cdot,\cdot]$ denotes channel-wise concatenation, \(\psi(\cdot)\) denotes a lightweight convolutional layer, and \(\sigma(\cdot)\) is the sigmoid function, and $\odot$ denotes element-wise multiplication.
The final output feature \(F\) serves as the enhanced BEV representation for subsequent collaborative fusion.

\subsection{Active Feature Communication}

To reduce redundant communication, we incorporate a Where2comm-style active communication mechanism~(\cite{hu2022where2comm}) into our framework. 
This module estimates the communication value of each BEV location from the enhanced BEV feature \(F^{(k)}\) and selectively transmits informative regions instead of the entire dense feature map. 
Specifically, a lightweight confidence generator predicts a spatial confidence map, which is thresholded to obtain a binary communication mask. 
The transmitted feature is then computed by applying this mask to \(F^{(k)}\). 
In our framework, this selection is performed on the DINOv2-guided enhanced BEV representation, and the resulting sparse feature is used for subsequent ego-centric collaborative fusion.

This active communication mechanism connects single-agent feature enhancement with cross-agent fusion. 
It determines which regions should be transmitted from each agent, while the subsequent cooperative fusion module focuses on how to incorporate the received complementary features into the ego representation.

% \begin{figure}[t]
%   \centering
%   \includegraphics[width=\linewidth]{image/ego_centric.png}
%   \caption{Ego-Centric Collaborative Feature Fusion.}
%   \label{fig:ego_centric}
% \end{figure}

\subsection{Ego-Centric Collaborative Fusion} \label{sec:ma_fusion}

After the ego vehicle receives selected features from neighboring agents, we first transform them into the ego coordinate system and then perform feature aggregation. We adopt an ego-centric residual fusion strategy, where the ego feature serves as the primary representation and neighboring features provide complementary residual information. Here, a gating mechanism is further introduced to adaptively adjust the information contribution ratios of ego and neighboring features in the corresponding spatial regions, leading to a more robust cross-agent fusion result.

Specifically, let \(F_{{ego}}\) denote the ego feature and \(\{F_{{nb}}^k\}_{k=1}^{K}\) denote the features received from neighboring agents after spatial alignment. We first align the ego feature and each neighboring feature with two independent projection functions:
\begin{equation}
\tilde{F}_{\text{ego}} = \phi_e(F_{\text{ego}}), \qquad
\tilde{F}_{\text{nb}}^{k} = \phi_n(F_{\text{nb}}^{k}), \quad k=1,\dots,K,
\end{equation}
where \(\phi_e(\cdot)\) and \(\phi_n(\cdot)\) denote learnable alignment mappings for the ego and neighbor branches, respectively.

After spatial alignment, we use a lightweight gating network to estimate where the ego feature may benefit from neighboring information. 
Specifically, the gate is predicted based on the ego feature, the neighboring feature, and their absolute difference:
\[
G_k = \sigma \left( \psi_g \left( [\tilde{F}_{ego}, \tilde{F}_{nb}^{k}, |\tilde{F}_{nb}^{k}-\tilde{F}_{ego}|] \right) \right),
\]
where \(\psi_g(\cdot)\) denotes a lightweight convolutional network, \(\sigma(\cdot)\) is the sigmoid function, and \([\cdot]\) denotes channel-wise concatenation. 
The gate \(G_k\) adaptively indicates the BEV regions where neighboring feature should contribute more to compensate for insufficient ego observations.

Finally, the collaborative fusion result is obtained by applying a residual update to the ego feature:
\begin{equation}
F_{{coop}} = F_{{ego}} + \alpha \odot R,
\end{equation}
where \(\alpha\) is a learnable channel-wise scaling parameter. In this way, the fused representation remains centered on the ego feature, while selectively incorporating complementary information from neighboring agents through spatially adaptive residual aggregation.
Finally, the output feature \(F_{coop}\) is fed into the downstream detection head for final 3D object detection.

\section{Experiments}

\begin{table}[t]
\centering
\captionsetup{skip=6pt}

\caption{Comparison with state-of-the-art methods on DAIR-V2X and V2XSet validation datasets.}
\label{tab:performace_compare}

\renewcommand{\arraystretch}{1.20}
\setlength{\tabcolsep}{5pt}
% 使用 >{\centering\arraybackslash}p{} 让第3、4列内容自动居中
\begin{tabular}{c|c|>{\centering\arraybackslash}p{3.6cm}|>{\centering\arraybackslash}p{3.6cm}}
% \toprule
\hline

\textbf{Fusion} 
& \textbf{Model} 
& \begin{tabular}[c]{@{}c@{}}\textbf{DAIR-V2X} \\ \textbf{AP@0.3/0.5/0.7}\end{tabular}
& \begin{tabular}[c]{@{}c@{}}\textbf{V2XSet} \\ \textbf{AP@0.3/0.5/0.7}\end{tabular} \\
\hline

\multirow{1}{*}{\textit{No Fusion}} 
& PointPillars (2019)
& 72.32\,/\,69.29\,/\,59.86
& 66.92\,/\,64.03\,/\,41.42 \\
\hline

\multirow{1}{*}{\textit{Late Fusion}} 
& PP-LF (2019)
& 79.24\,/\,67.76\,/\,49.92
& 86.30\,/\,83.72\,/\,58.25 \\
\hline

\multirow{6}{*}{\begin{tabular}[c]{@{}c@{}}\textit{Intermediate}\\ \textit{Fusion}\end{tabular}}
& PP-IF (2019)
& 75.87\,/\,68.31\,/\,51.98
& 95.28\,/\,89.67\,/\,65.75 \\

& Where2comm (2022)
& 82.65\,/\,\underline{78.94}\,/\,64.73
& 87.11\,/\,84.41\,/\,71.61 \\

& CoAlign (2023)
& 82.65\,/\,77.55\,/\,62.64
& \underline{96.68}\,/\,\underline{95.83}\,/\,\underline{88.86} \\

& DI-V2X (2024)
& 81.58\,/\,77.77\,/\,65.54
& 94.96\,/\,93.93\,/\,85.14 \\

& INSTINCT (2025)
& \underline{82.76}\,/\,78.44\,/\,\underline{69.39}
& 95.61\,/\,94.59\,/\,87.97 \\

\cline{2-4}
& Ours
& \textbf{86.35\,/\,82.80\,/\,71.39}
& \textbf{97.91\,/\,97.21\,/\,93.25} \\

% \bottomrule
\hline

\end{tabular}
\end{table}

To comprehensively evaluate \systemname, we conduct experiments on two widely adopted benchmarks (Dair-V2X and V2XSim) for collaborative perception. This section presents the experimental setup, quantitative results, and analysis. Due to space limitations, detailed descriptions of the datasets and evaluation metrics are provided in the supplementary material~(App.\ref{sec:supp_mater}).

\subsection{Experimental Setup}
\label{sec:exp_setup}

\subsubsection{Implementation Details}
We set the point cloud range to \([-100.8, 100.8] \times [-44.8, 44.8] \times [-3.5, 1.5]\)~m in the vehicle coordinate system, with the voxel size set to \([0.4, 0.4, 5]\)~m along the \(X\), \(Y\), and \(Z\) axes, respectively. To preserve more spatial details, we use a planar resolution of \(0.1\)~m for BEV projection. In the DINOv2 branch, we adopt \texttt{dinov2\_vits14} as the pretrained visual backbone, and freeze its first eight blocks during training. 

To ensure a fair comparison, all methods are evaluated under a unified BEV perception setting. We note that different methods may adopt different detection heads, and some of them are not implemented based on the PointPillars detector. Therefore, instead of enforcing a strictly unified detector architecture, we compare all methods under the same input range and a consistent BEV representation setting.

\subsection{Main Result}
% \input{table/compare_with_sort}
% 表1展示了多种V2X 方法的对比，包括早期融合、中间融合和后期融合方法。其中，PP 表示使用车辆端和路端混合数据共同训练的 PointPillars。可以观察到，后融合方法PP-LF在V2XSet中取得了提升，然而在DAIR-V2x上，在AP@0.3指标上有所提升，但是在AP@0.5以及AP@0.7上有所下降，这主要是真实场景中存在的位姿误差导致的。朴素的特征融合策略要优于后融合的策略。
Tab.~\ref{tab:performace_compare} presents the comparison of different V2X collaborative perception methods, including intermediate-fusion, and late-fusion approaches. Here, PP denotes PointPillars trained jointly with mixed vehicle-side and roadside data. It can be observed that the late-fusion method PP-LF achieves performance gains on V2XSet. However, on DAIR-V2X, although PP-LF improves AP@0.3, its performance drops on AP@0.5 and AP@0.7 due to pose errors in real-world scenarios. In contrast, the naive feature-fusion strategy outperforms the late-fusion one.

% 此外，我们还将方法与一些具有竞争力的中间融合方法进行了比较。INSTINCT复现了其基于BEV检测的版本，在DAIR-V2X数据集上，表现均优于DI-V2X。可以看到，我们的方法相比于现有的模型取得了进一步的提升。我们认为这主要是因为我们提取了更多样地特征，从而更好地发挥了多视角地补充作用，后续我们进行更细节地讨论。在附录中，我们也给出了一些可视化的结果。
Additionally, we compare our method with several competitive intermediate-fusion methods. INSTINCT~(\cite{xu2025instinct}) reproduces its BEV-detection-based version, which outperforms DI-V2X~(\cite{li2024di}) on the DAIR-V2X dataset. It can be observed that our method achieves further improvements over existing models. 

On the V2XSet, our method also achieves the best performance across all IoU thresholds. The improvement is particularly significant under the stricter IoU threshold of 0.7, indicating that our method not only enhances object detection recall but also produces more accurate localization results. 

These results demonstrate the effectiveness of our method in leveraging complementary multi-view information for collaborative detection.
We primarily attribute this to the fact that our method extracts more diverse features, thereby better exploiting the complementary information from multiple viewpoints. We provide a more detailed discussion in the following sections. Additional qualitative visualization results are also provided in App.~\ref{sec:supp_mater}.

\subsection{Gain Comparison}
\begin{table}[t]
\centering
\captionsetup{skip=6pt}

\caption{Comparison between \textbf{baselines} and \textbf{ours} under single and fusion settings.}
% \resizebox{\linewidth}{!}{
\begin{tabular}{l l c c c}
\toprule
\textbf{Method} & \textbf{Setting} & \textbf{AP@0.3} & \textbf{AP@0.5} & \textbf{AP@0.7} \\
\midrule
\multirow{2}{*}{Where2comm}
& Single & 71.60 & 68.03 & 57.49 \\
& Fusion & \textbf{82.65} {\color{red}(+11.05)} & \textbf{78.94} {\color{red}(+10.91)} & \textbf{64.73} {\color{red}(+7.24)} \\
\midrule
\multirow{2}{*}{INSTINCT}
& Single & 75.26 & 71.57 & 63.70 \\
& Fusion & \textbf{82.76} {\color{red}(+7.50)} & \textbf{78.44} {\color{red}(+6.87)} & \textbf{69.39} {\color{red}(+5.69)} \\
\midrule
\multirow{2}{*}{DI-V2X}
& Single & 73.73 & 69.77 & 59.22 \\
& Fusion & \textbf{81.58} {\color{red}(+8.25)} & \textbf{77.88} {\color{red}(+8.11)} & \textbf{65.54} {\color{red}(+6.32)} \\
\midrule
\multirow{2}{*}{Ours}
& Single & 72.70 & 69.79 & 62.65 \\
& Fusion & \textbf{86.35} {\color{red}\textbf{(+13.65)}} & \textbf{82.80} {\color{red}\textbf{(+13.01)}} & \textbf{71.39} {\color{red}\textbf{(+8.74)}} \\
\bottomrule
\end{tabular}
% }
\label{tab:gain_comapare}
\end{table}

To demonstrate that our method extracts diversity features and better facilitate collaborative perception, we compare the performance of our method with existing baselines under both single-agent and fusion settings. 
We further report the performance gains brought by collaboration for each method. 
As shown in Tab.~\ref{tab:gain_comapare}, all methods benefit from collaborative fusion, while our method achieves the largest improvements across all IoU thresholds. 
Specifically, compared with the single-agent setting, our method improves AP by \(13.65\%\), \(13.01\%\), and \(8.74\%\) at IoU thresholds of 0.3, 0.5, and 0.7, respectively. 
These gains outperform those of Where2comm, INSTINCT, and DI-V2X, indicating that the features learned by our method are more suitable for cross-agent alignment and fusion. 
In the next subsection, we further analyze this improvement from the perspective of the feature space.

\subsection{Feature Space Analysis: Alignment vs. Complementarity}

To address the data distribution discrepancies among different agents in V2X collaborative perception, existing methods often encourage agents to learn highly consistent feature representations for easier alignment and fusion. 
However, due to differences in sensor positions, viewpoints, and visible regions, different agents tend to focus on different aspects of the scene, and their feature representations should preserve such complementary differences. 
Simply pursuing cross-agent feature consistency may suppress the complementary information among agents, thereby limiting the benefit of multi-view collaboration.

\begin{figure}[t]
    \centering
    \begin{minipage}[t]{0.49\linewidth}
        \centering
        \includegraphics[width=\linewidth]{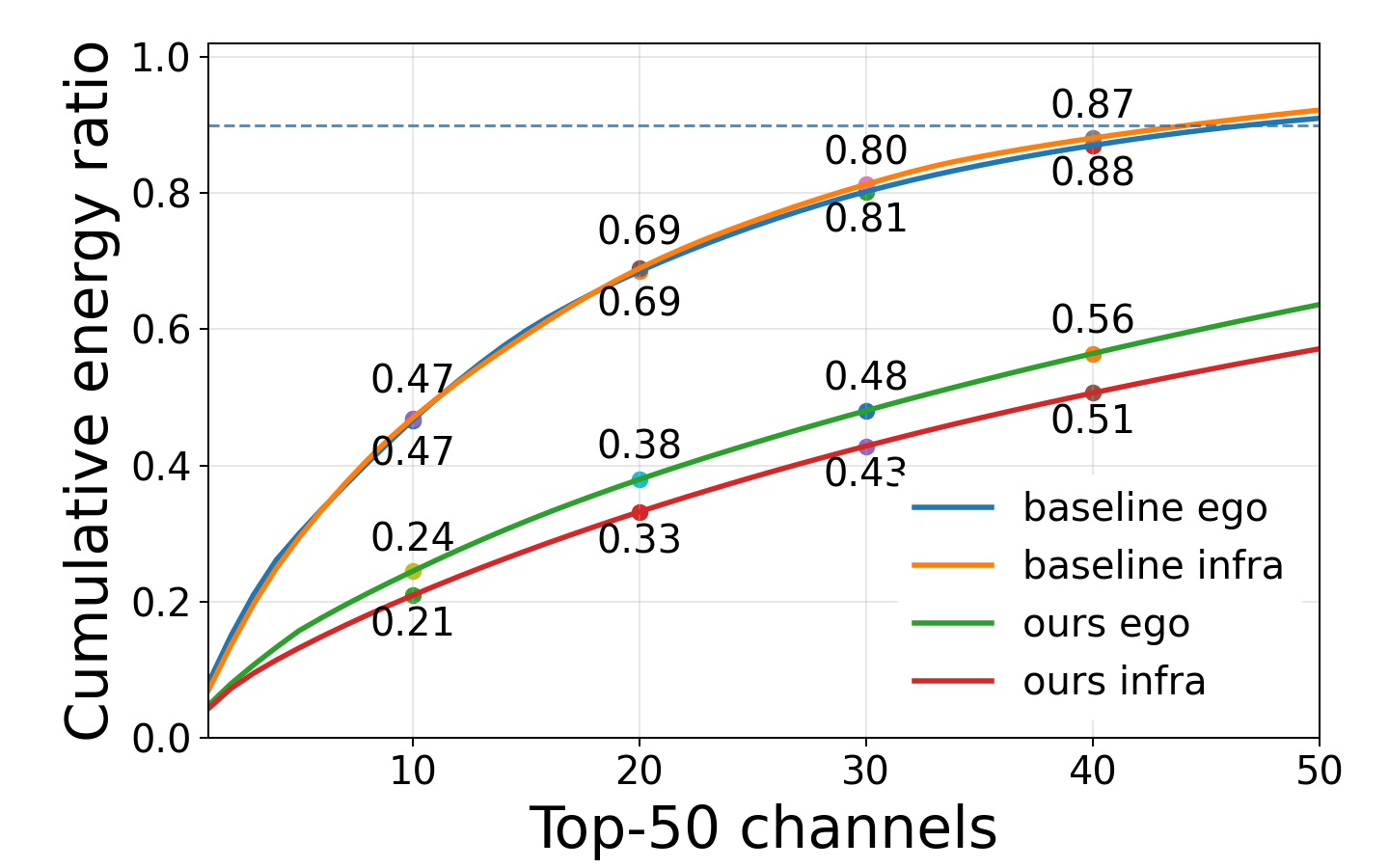}
        \caption{Cumulative channel energy ratio.}
        \label{fig:energy}
    \end{minipage}
    \hfill
    \begin{minipage}[t]{0.49\linewidth}
        \centering
        \includegraphics[width=\linewidth]{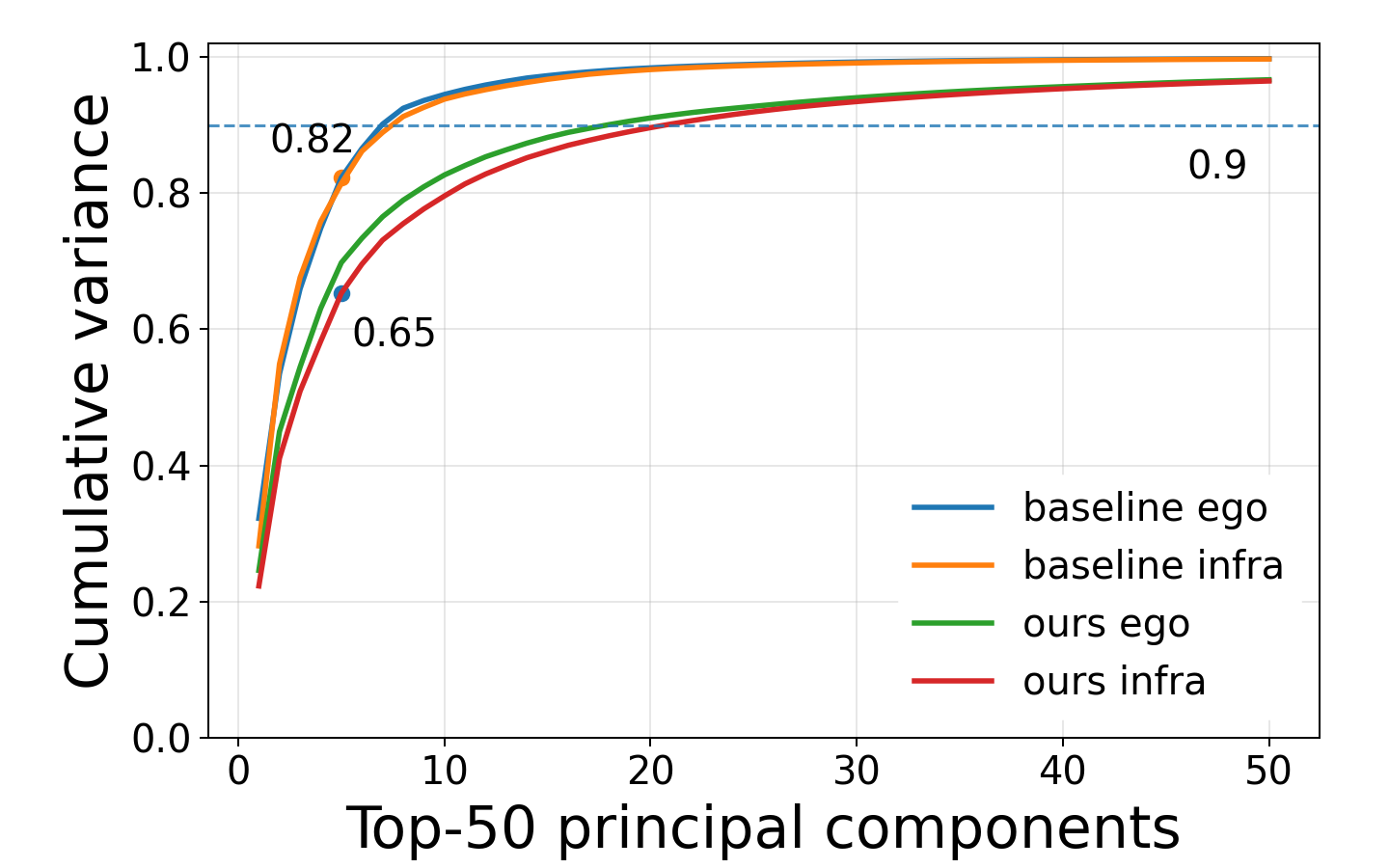}
        \caption{PCA cumulative explained variance.}
        \label{fig:pca}
    \end{minipage}
    \label{fig:energy_pca_side_by_side}
\end{figure}

First, we sort the channel-wise energy in descending order and compute the cumulative energy ratio. As shown in Fig.~\ref{fig:energy}, the baseline features are highly concentrated in a small number of dominant channels. For example, the top-40 channels already account for about \(0.87\) and \(0.88\) of the total energy for ego and infrastructure features, respectively. In contrast, \systemname distributes feature energy more evenly across channels, where the top-40 channels only account for about \(0.56\) and \(0.51\). This indicates that \systemname activates a broader set of channels for feature representation, rather than relying on a few dominant channels.

Second, we perform PCA analysis on the foreground features. As shown in Fig.~\ref{fig:pca}, the first five principal components of the baseline explain about \(0.82\) of the variance, while those of \systemname explain only about \(0.65\), indicating that the baseline concentrates approximately \(1.26\times\) more variance in the leading components. 
Moreover, the baseline reaches a cumulative explained variance of \(0.9\) with only around seven principal components, whereas \systemname requires more than twenty components to reach a comparable level. This suggests that the foreground representation learned by \systemname occupies a more distributed feature subspace.

These observations indicate that the DINOv2-guided representation does not simply enforce feature similarity across agents. Instead, it constructs a richer and more diversity feature space, allowing different agents to preserve complementary information from their own observations. This property helps explain why \systemname achieves larger gains under collaborative fusion.

\subsection{Ablation Studies}

\subsubsection{Ablation on PP-DINOv2 Feature Fusion}

\begin{table*}
\centering
\begin{minipage}[t]{0.40\textwidth}
    \centering
    % \begin{table}[t]
\centering
\caption{Ablation of feature branch.}
% \resizebox{0.9\linewidth}{!}{
\setlength{\tabcolsep}{4pt}
\begin{tabular}{lccc}
\toprule
% Metric & Only-PP & Only-DINO & PP+DINO \\
Metric &
\shortstack{Only-\\PP} &
\shortstack{Only-\\DINOv2} &
\shortstack{PP+\\DINOv2} \\
% \textbf{Metric} & \textbf{Only-PP} & \textbf{Only-DINO} & \textbf{PP+DINO} \\
% \textbf{Metric} & \textbf{Only-\\PP} & \textbf{Only-\\DINO} & \textbf{PP+\\DINO} \\
\midrule
AP@0.3   & 85.91 & 35.92 & \textbf{86.35} \\
AP@0.5   & 81.56 & 14.49 & \textbf{82.80} \\
AP@0.7   & 68.52 & 1.74  & \textbf{71.39} \\
Recall    & 73.99 & 7.72  & \textbf{75.56} \\
Precision & 62.55 & 10.11 & \textbf{65.85} \\
\bottomrule
\end{tabular}
% }
\label{tab:diff_branch}
% \end{table}
\end{minipage}
\hfill
\begin{minipage}[t]{0.58\textwidth}
    \centering
    % \begin{table}[t]
\centering
\caption{Ablation study of the proposed framework.}
\vspace{1em}
\label{tab:ablation_main}
\setlength{\tabcolsep}{3pt}
\renewcommand{\arraystretch}{1.15}
\begin{tabular}{ccc|cccc}
\hline
 LRF & HRF & GRF & AP@0.3 & AP@0.7 & Recall & Precision \\
\hline
-- & -- & -- & 82.65 & 64.71 & 69.95 & 63.96 \\
\hline
\checkmark & -- & -- & 85.00 & 65.96 & 72.34 & 58.56 \\
\checkmark & \checkmark & -- & 85.97 & 71.02 & 75.31 & \textbf{66.20} \\
\checkmark & \checkmark & \checkmark & \textbf{86.48} & \textbf{71.42} & \textbf{75.69} & 65.12 \\
\hline
\end{tabular}
% \end{table}
\end{minipage}
\end{table*}

% 在 Section II 中，我们采用了以 PP 特征为主、DINO 特征为辅 的特征融合范式。为验证这一设计的合理性，本小节对特征分支进行了局部消融实验。在固定其余模块与训练配置的前提下，我们比较了三种设置：仅使用 PP 特征、仅使用 DINO 特征，以及 PP 为主、DINO 为辅的融合方式。结果如表 II 所示。可以看到，仅使用 PP 特征时已经能够取得较强的检测性能，而仅使用 DINO 特征时性能明显较弱；进一步结合 DINO 后，整体结果在各项指标上均优于仅使用 PP 特征的设置。这说明 PP 特征能够作为检测任务的主体表示，而 DINO 虽然不适合作为独立的检测主分支，但能够提供额外的互补信息，从而有效增强 PP 特征。该实验验证了我们所提出特征分支设计的合理性。我们认为，这主要是因为 PP 特征能够较好地保留局部几何结构与空间细节信息，这些信息对于目标检测任务尤其关键。尽管如此，在点云较为稀疏或目标存在部分遮挡时，这类局部几何信息往往不够完整，因而会限制检测性能。相较而言，DINO 特征更侧重于高层语义表示，能够在此基础上提供额外的语义上下文与鲁棒性补充，因此更适合作为辅助分支对 PP 特征进行补充。

In Sec.~\ref{sec:feature_fusion}, we adopt a fusion design that combines PP features with DINOv2 features.
To validate this design, we compare three configurations while keeping other modules unchanged: PP-only, DINOv2-only, and PP-DINOv2 fusion. 
As shown in Tab.~\ref{tab:diff_branch}, PP-only already achieves strong detection performance, whereas DINOv2-only performs much worse, indicating that DINOv2 features are not suitable as standalone detection features. 
However, incorporating DINOv2 with Pillar-branch consistently improves all metrics. 
This experiment validates the effectiveness of our design.

\subsubsection{Ablation on Different Modules}

As shown in Tab.~\ref{tab:ablation_main}, the proposed components bring consistent improvements. 
Low-resolution fusion benefits global context modeling and target discovery, but using it alone may introduce additional false positives. 
High-resolution refinement alleviates this issue by enhancing local details and improving localization. 
The collaborative fusion module further incorporates complementary information from neighboring agents, resulting in the best overall performance.

% 车辆数量变化对于性能的影响 ？
% \subsubsection{Compare different number of car}

% 测试不同的dino_v2的版本，
% \subsubsection{DINOv2 variants: what changes across versions}

% 测试不同的 pp 的特征维度
% \subsubsection{Compare different PP feature dimentation}

\section{Conclusion And Discussion}
\label{sec:conclusion}

This paper presents \systemname, a novel framework for vehicle-infrastructure collaborative perception. By introducing vision foundation models into point cloud-based pipelines, the proposed framework effectively enhances feature representation and improves perception performance. Compared with existing methods, our method achieves not only the best collaborative perception results but also the largest collaborative gain. 

% These improvements may contribute to more reliable perception in autonomous driving systems, potentially enhancing safety and traffic efficiency in real-world scenarios. 
% While achieving strong performance, the method is currently evaluated on only two V2X benchmarks, and its robustness under challenging conditions such as severe sensor noise and extreme data sparsity remains to be further explored.
% Despite promising performance, our method is evaluated on only two V2X benchmarks, and its robustness under challenging conditions remains to be further explored.
% In the future, we aim to further explore the potential of larger vision foundation models and extend the proposed framework to more challenging settings, such as localization noise, communication delays, and multimodal sensor inputs.

{
\small
\bibliographystyle{plainnat}
\bibliography{references}
}
%%%%%%%%%%%%%%%%%%%%%%%%%%%%%%%%%%%%%%%%%%%%%%%%%%%%%%%%%%%%
\newpage

\appendix

\section{Technical appendices and supplementary material}
\label{sec:supp_mater}

\subsection{Memory Cost Analysis of Attention-Based Fusion}
\label{sec:appendix_memorycost}

\begin{table}
\centering
\captionsetup{skip=6pt}
\caption{Comparison of feature scales under different downsampling ratios.}
\label{tab:downsample_compare}
\setlength{\tabcolsep}{5pt}
\begin{tabular}{lccc}
\toprule
\textbf{Metric} & \textbf{Original} & \textbf{\(2\times\)} & \textbf{\(4\times\)} \\
\midrule
Spatial size & \(112 \times 252\) & \(56 \times 126\) & \(28 \times 63\) \\
Tokens / modality & \(28{,}224\) & \(7{,}056\) & \(1{,}764\) \\
Query tokens & \(28{,}224\) & \(7{,}056\) & \(1{,}764\) \\
Memory tokens & \(56{,}448\) & \(14{,}112\) & \(3{,}528\) \\
Token reduction & \(1\times\) & \(4\times\) & \(16\times\) \\
Attention cost & \(1\times\) & \(1/16\) & \(1/256\) \\
Attention score (FP32) & \(48.62\) GB & \(2.07\) GB & \(0.186\) GB \\
\bottomrule
\end{tabular}
\label{tab:downsample}
\end{table}

Tab.~\ref{tab:downsample} analyzes the computational cost of applying attention-based fusion under different BEV feature resolutions. 
At the original resolution of \(112 \times 252\), each branch contains \(28{,}224\) tokens, resulting in \(56{,}448\) memory tokens for cross-feature interaction. 
The corresponding attention score matrix requires \(48.62\) GB of memory in FP32, making direct attention-based fusion on dense BEV features impractical. 

Downsampling effectively reduces the token number and attention cost. 
With a \(2\times\) downsampling ratio, the attention cost is reduced to \(1/16\) of the original cost, and the FP32 attention memory decreases to \(2.07\) GB. 
When using a \(4\times\) downsampling ratio, the attention cost is further reduced to \(1/256\), requiring only \(0.186\) GB of memory. 
Therefore, we perform global attention-based fusion in the \(4\times\) downsampled space, which substantially reduces the memory overhead while still allowing long-range interaction between PointPillars and DINOv2 features.

\subsection{Experimental Datasets}
\label{sec:dataset}

\noindent\textbf{DAIR-V2X.} We evaluate our method on the challenging {DAIR-V2X} dataset~(\cite{yu2022dair}), a large-scale benchmark for vehicle-infrastructure cooperative perception in real-world scenarios. The dataset contains around 9,000 synchronized vehicle and infrastructure LiDAR frames from 100 representative scenes at a frequency of 10~Hz. The roadside unit is equipped with a 300-line solid-state LiDAR, while the vehicle is equipped with a 40-line mechanical LiDAR.

\noindent\textbf{V2XSet.} We further evaluate our method on the {V2XSet} dataset~(\cite{xu2022v2x}), which explicitly models realistic V2X noise. The dataset consists of 6,694 training samples and 1,920 validation samples. Each scene includes point clouds collected from 2 to 7 agents, which are equipped with 36-beam LiDAR sensors, providing a 360$^\circ$ horizontal field of view.

\subsection{Licenses for Existing Assets}
\label{appendix:license}

This work uses publicly available datasets and pretrained models. 
DAIR-V2X\footnote{\url{https://github.com/AIR-THU/DAIR-V2X}} is released under the Apache 2.0 license. 
V2XSet\footnote{\url{https://github.com/DerrickXuNu/v2x-vit}} is used under its research usage terms. 
DINOv2\footnote{\url{https://github.com/facebookresearch/dinov2}} is released under the Apache 2.0 license by Meta AI.

All assets are properly cited in the main paper, and we adhere to their respective terms of use.

\subsection{Compute Resources}
\label{sec:compute_resources}
% \answerTODO{}
% All experiments are conducted on NVIDIA GPUs (e.g., A100 or RTX 3090). Each model is trained using 4 GPUs, with an approximate training time of 12--24 hours per experiment depending on the configuration.
All main experiments are conducted on the DAIR-V2X dataset using a server equipped with four NVIDIA RTX A6000 GPUs (48GB memory each) and a 32-core CPU with 32 data loading workers.. 
We adopt a multi-GPU data-parallel training strategy to accelerate training and improve throughput. 
Under this setting, a single training epoch on DAIR-V2X takes approximately 30 to 40 minutes, and the full training process typically completes within 90 epochs. This setup enables efficient experimentation and ablation studies while maintaining consistent and fair training conditions across all compared methods.

\subsection{Computation Metric Details}

A predicted bounding box is considered a true positive (TP) if its Intersection-over-Union (IoU) with a ground-truth box exceeds a predefined threshold and the predicted category is correct. Otherwise, it is counted as a false positive (FP). 
A ground-truth object that is not matched by any prediction is counted as a false negative (FN).

The IoU between a predicted box \(B_p\) and a ground-truth box \(B_g\) is defined as:

\begin{equation}
\text{IoU} = \frac{|B_p \cap B_g|}{|B_p \cup B_g|}
\end{equation}

where \(|B_p \cap B_g|\) and \(|B_p \cup B_g|\) denote the intersection and union areas (or volumes) of the two boxes, respectively.

A higher Precision indicates that the model produces fewer incorrect detections, meaning that most predicted objects are correct; however, it does not reflect missed detections. In contrast, a higher Recall indicates that the model misses fewer ground-truth objects, meaning that more real objects are successfully detected, but it may also introduce more false positives and thus reduce Precision. Therefore, relying on either metric alone cannot fully reflect the overall detection performance. To jointly evaluate both Precision and Recall, Average Precision (AP) is introduced as the area under the Precision--Recall (PR) curve.

Precision and Recall are defined as:
\begin{equation}
\text{Precision} = \frac{TP}{TP + FP}, \quad
\text{Recall} = \frac{TP}{TP + FN}.
\end{equation}
Where \(TP\), \(FP\), and \(FN\) denote the numbers of true positive, false positive, and false negative detections, respectively.

Average Precision (AP) is computed as the area under the Precision–Recall curve:
\begin{equation}
\text{AP} = \int_0^1 p(r)\, dr,
\end{equation}
Where \(p(r)\) represents the precision value at recall level \(r\).
We report AP at IoU thresholds of 0.30, 0.50, and 0.70 denoted as AP@0.3, AP@0.5, and AP@0.7.

\subsection{Visualization Results}

\begin{figure*}[t]
    \centering
    \includegraphics[width=1\linewidth]{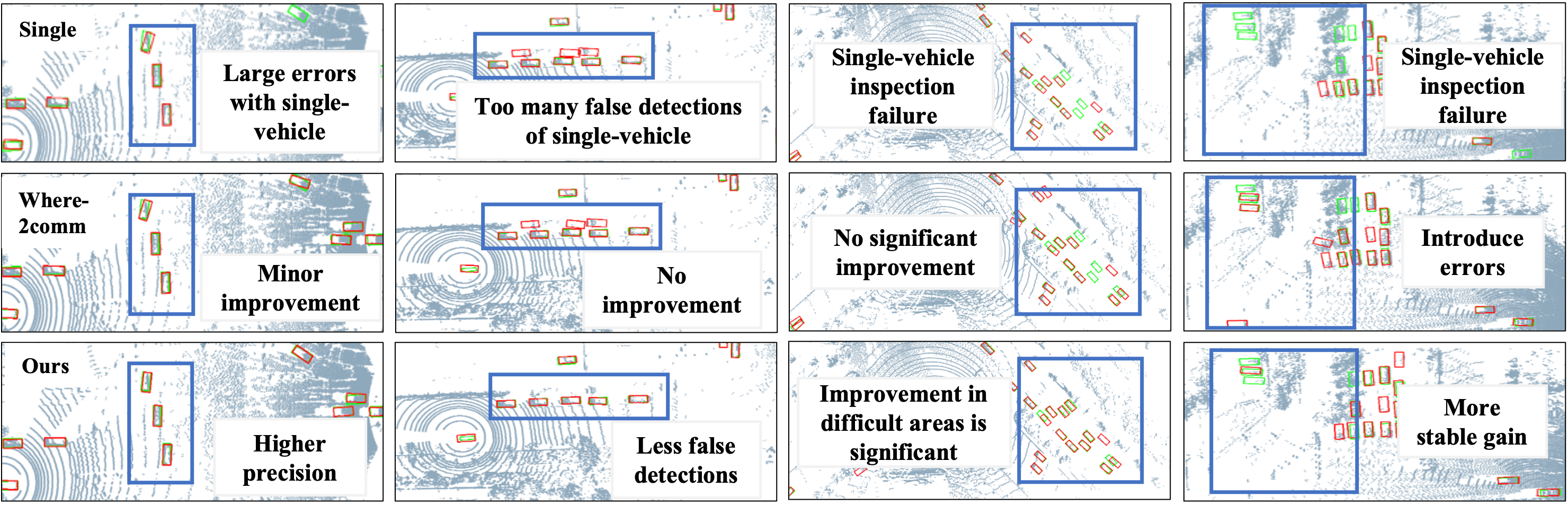}
    \caption{Visualization Results Comparison.}
    \label{fig:det_figure}
\end{figure*}

We provide qualitative visualization results in the supplementary material to further demonstrate the effectiveness of \systemname. 
Specifically, we compare Single, Where2comm, and our method under multiple representative scenarios. 
As shown in Fig.~\ref{fig:det_figure}, \systemname achieves more accurate detection results with fewer false positives. 
In addition, our method shows improved detection performance in challenging regions, where objects are difficult to perceive due to sparse observations, occlusions, or limited viewpoints. 
These results further indicate that the proposed DINOv2-guided feature learning and multi-scale fusion 
strategy can enhance collaborative perception in complex driving scenes.

\subsection{Collaborative Gain Definition}

The goal of collaborative perception is not only to achieve high detection performance, but also to effectively exploit complementary observations from other agents. 
However, the absolute performance under the collaborative setting cannot fully reflect how much a method benefits from collaboration. 
For example, a method may achieve high collaborative performance mainly because its single-agent detector is strong, while the additional improvement brought by cross-agent fusion is limited. 
Therefore, to explicitly measure the benefit brought by collaboration itself, we introduce the collaborative gain metric. Formally, it is computed as
\[
\Delta M = M_{fusion} - M_{single},
\]
where \(M_{single}\) denotes the performance obtained using only the ego agent's observation, \(M_{fusion}\) denotes the performance obtained after collaborative fusion, and \(M\) can be any evaluation metric, such as AP at a specific IoU threshold.
 
A larger \( \Delta M \) indicates that the model obtains more benefit from cross-agent collaboration. This metric is important because a larger collaborative perception gain suggests that the learned feature representation and fusion mechanism are more favorable for cross-agent information sharing. Therefore, in this paper, we report collaborative perception gains under different IoU thresholds to evaluate how much different methods benefit from collaboration.

\subsection{ Regional Similarity Evaluation and Analysis }

\begin{table}[t]
\centering
\captionsetup{skip=6pt}
\caption{Extended cross-view feature similarity comparison.}
\label{tab:region_similarity_extended}
\begin{tabular}{llccc}
\toprule
Metric & Region & Baseline & Ours & $\Delta$ (Ours$-$Base) \\
\midrule
Cosine      & All      & 0.8243 & 0.6622 & -0.1621 \\
Cosine      & FG-Both  & 0.8316 & 0.5140 & -0.3176 \\
Cosine      & BG-Both  & 0.8257 & 0.6742 & -0.1515 \\
Cosine      & FG-Union & 0.7537 & 0.4858 & -0.2680 \\
\midrule
Correlation & All      & 0.8055 & 0.5941 & -0.2114 \\
Correlation & FG-Both  & 0.8086 & 0.4085 & -0.4001 \\
Correlation & BG-Both  & 0.8072 & 0.6090 & -0.1983 \\
Correlation & FG-Union & 0.7229 & 0.3768 & -0.3461 \\
\bottomrule
\end{tabular}
\end{table}
\begin{table}
\centering
\captionsetup{skip=6pt}
\caption{Ablation study on the number of unfrozen DINO blocks during fine-tuning.}
\label{tab:dino_unfreeze}
\setlength{\tabcolsep}{4pt}
\renewcommand{\arraystretch}{1.15}
\begin{tabular}{c|ccc}
\hline
% \toprule
% Unfrozen \\ DINO blocks & AP@0.3 & AP@0.5 & AP@0.7 \\
\makecell{Unfrozen \\ DINO blocks} &
\makecell{AP \\ IoU@0.3} &
\makecell{AP \\ IoU@0.5} &
\makecell{AP \\ IoU@0.7} \\
\hline
0  & 85.93 & 81.33 & 68.02 \\
4  & \textbf{86.35} & \textbf{82.80} & \textbf{71.39} \\
12 & 73.77 & 70.54 & 63.03 \\
\hline
% \bottomrule
\end{tabular}
\end{table}

To quantify the consistency between vehicle-side and infrastructure-side representations, we compute cross-view feature similarity in the BEV feature space. 
Let the BEV features from the vehicle and infrastructure branches be denoted by
\[
\mathbf{F}^{v}, \mathbf{F}^{i} \in \mathbb{R}^{C \times H \times W},
\]
where \(C\), \(H\), and \(W\) denote the channel number, height, and width, respectively. 
For each BEV location \((u,v)\), we extract the corresponding channel vectors
\[
\mathbf{f}^{v}_{u,v} \in \mathbb{R}^{C}, \qquad \mathbf{f}^{i}_{u,v} \in \mathbb{R}^{C},
\]
and compute their cosine similarity as
\[
s_{u,v}
=
\frac{\mathbf{f}^{v}_{u,v} \cdot \mathbf{f}^{i}_{u,v}}
{\|\mathbf{f}^{v}_{u,v}\|_2 \, \|\mathbf{f}^{i}_{u,v}\|_2 + \epsilon},
\]
where \(\epsilon\) is a small constant for numerical stability. 
This yields a spatial similarity map
\[
\mathbf{S} \in \mathbb{R}^{H \times W}.
\]

To avoid the influence of invalid spatial positions, we only evaluate similarity over locations where both branches have valid feature responses. 
Furthermore, to distinguish the behavior in different semantic regions, we construct foreground masks from the anchor labels and report the average cosine similarity over three regions: 
(1) all valid locations (All), 
(2) the intersection of foreground regions from both views (FG-Both), and 
(3) the union of foreground regions from the two views (FG-Union). 
Specifically, if \(\mathbf{M}^{v}_{fg}\) and \(\mathbf{M}^{i}_{fg}\) denote the foreground masks of the vehicle and infrastructure branches, respectively, then
\[
\mathbf{M}_{\text{FG-Both}} = \mathbf{M}^{v}_{fg} \cap \mathbf{M}^{i}_{fg},
\qquad
\mathbf{M}_{\text{FG-Union}} = \mathbf{M}^{v}_{fg} \cup \mathbf{M}^{i}_{fg}.
\]
The reported similarity is obtained by averaging \(\mathbf{S}\) over the corresponding masked region. 
A higher value indicates stronger cross-view feature consistency. 

Tab.~\ref{tab:region_similarity_extended} reports the average cosine similarity in the above three regions, where a larger value indicates higher cross-view feature consistency. Although the proposed method achieves clear improvements in the final detection performance, the cross-view cosine similarity reported in Tab.~\ref{tab:region_similarity_extended} is consistently lower than the baseline in the All, FG-Both, and FG-Union regions. This indicates that our method does not explicitly increase the consistency between vehicle-side and infrastructure-side features. We believe this is mainly because our optimization objective is not cross-view feature alignment, but rather the learning of more discriminative task-oriented representations and more effective cooperative fusion. In vehicle–infrastructure cooperation, preserving a certain degree of cross-view discrepancy can itself be beneficial, since such discrepancy may provide complementary information for subsequent fusion. Combined with the ablation results, it can be observed that the DINOv2 branch in our framework mainly serves as an auxiliary semantic branch. Its value lies in enhancing single-agent representations and improving their utility for downstream cooperative fusion, rather than explicitly pulling the cross-view features closer in the shared space. Therefore, the decrease in cosine similarity is not necessarily inconsistent with the observed performance gains; instead, it suggests that the proposed method tends to learn complementary task-driven representations rather than simply more aligned ones.

% 测试lora的不同层冻结的情况
\subsection{Ablation on the number of LoRA-adapted DINOv2 blocks}

As shown in Tab.~\ref{tab:dino_unfreeze}, partially fine-tuning the top layers of DINOv2 brings consistent improvements over the fully frozen setting. 
When the number of unfrozen blocks increases from 0 to 4, AP@0.3, AP@0.5, and AP@0.7 all improve, with the most notable gain observed on AP@0.7. 
This suggests that moderate adaptation of high-level semantic features is beneficial for the target detection task. 
However, when the number of unfrozen blocks is further increased to 12, the performance drops significantly on all metrics. 

A possible explanation is two-fold. 
First, the current task setting may not provide sufficient data scale and supervision strength to stably fine-tune a large portion of the pretrained DINOv2 backbone, and excessive updating may therefore weaken its pretrained visual priors. 
Second, in our framework DINOv2 mainly serves as an auxiliary semantic branch rather than a primary geometric detection backbone. 
Its main role is to provide high-level semantic context complementary to the PointPillars branch. 
Therefore, adapting only the top layers is more suitable, since it allows task-specific refinement of high-level semantics while preserving the general pretrained representation in the lower and middle layers. 
Overall, partially fine-tuning the top layers provides a better balance between task adaptation and prior preservation under the current setting.

\subsection{Feature-space metrics}

Fig.~\ref{fig:energy} and Fig.~\ref{fig:pca} show the cumulative channel-energy ratio and the cumulative PCA explained variance of the learned representations, respectively. Here we provide additional details on how these two quantities are computed, and further discuss what these results imply about the structure of the learned feature space.

\subsection{Channel Energy}

For a feature map \(F \in \mathbb{R}^{C \times H \times W}\), where \(C\) is the number of channels and \(H \times W\) is the spatial resolution, we define the energy of channel \(c\) as
\begin{equation}
E_c = \sum_{h=1}^{H}\sum_{w=1}^{W} F_{c,h,w}^2.
\end{equation}
This quantity measures the overall response magnitude of the \(c\)-th channel over the entire feature map. After computing \(\{E_c\}_{c=1}^{C}\), we sort the channel energies in descending order,
\begin{equation}
E_{(1)} \ge E_{(2)} \ge \cdots \ge E_{(C)},
\end{equation}
and define the cumulative energy ratio of the top-\(k\) channels as
\begin{equation}
R_k = \frac{\sum_{i=1}^{k} E_{(i)}}{\sum_{j=1}^{C} E_{(j)}}.
\end{equation}
The curve in Fig.~\ref{fig:energy} plots \(R_k\) as a function of \(k\). A rapidly increasing curve indicates that most of the response energy is concentrated in a small number of head channels, whereas a slower increase suggests that the energy is distributed across a broader set of channels.

\subsubsection{}{PCA-based variance analysis.}
To further analyze the feature distribution from the subspace perspective, we perform PCA on pooled feature vectors. Specifically, for each sample, we extract region-level averaged features such as global pooled features, foreground-intersection pooled features (FG-Both), and foreground-union pooled features (FG-Union). For a given feature type, these pooled vectors form a matrix
\begin{equation}
X \in \mathbb{R}^{N \times C},
\end{equation}
where \(N\) is the number of samples and \(C\) is the feature dimension. After mean-centering \(X\), PCA is applied to obtain eigenvalues
\begin{equation}
\lambda_1 \ge \lambda_2 \ge \cdots \ge \lambda_C.
\end{equation}
The explained variance ratio of the \(i\)-th principal component is
\begin{equation}
r_i = \frac{\lambda_i}{\sum_{j=1}^{C} \lambda_j},
\end{equation}
and the cumulative explained variance of the top-\(k\) principal components is
\begin{equation}
\mathrm{CEV}(k) = \frac{\sum_{i=1}^{k} \lambda_i}{\sum_{j=1}^{C} \lambda_j}.
\end{equation}
The curve in Fig.~\ref{fig:pca} plots \(\mathrm{CEV}(k)\) as a function of \(k\). A rapidly increasing PCA curve indicates that the feature variance is concentrated in a few dominant principal directions, while a slower curve indicates that the representation occupies a broader feature subspace and requires more principal components to explain the same amount of variance.

\subsubsection{Discussion.}
The two analyses characterize two complementary aspects of feature concentration. The channel-energy analysis is performed in the \emph{original channel basis}, and therefore reveals whether the representation is dominated by a small set of high-energy channels. In contrast, PCA is performed in the \emph{principal-direction basis}, and therefore reveals whether the feature variance is concentrated in a few dominant modes after a linear change of basis.

As shown in Fig.~\ref{fig:energy}, the baseline accumulates channel energy substantially faster than our method. For example, the top-\(10\) channels already account for a much larger portion of the total energy in the baseline, while our method requires many more channels to reach the same cumulative ratio. This indicates that the baseline representation is more strongly dominated by a few head channels, whereas our representation distributes activation energy across a broader set of channels.

Fig.~\ref{fig:pca} shows a consistent trend from the PCA perspective. The cumulative explained variance of the baseline rises much more rapidly than that of our method, especially on foreground features. In other words, the baseline can be explained by a relatively small number of dominant principal components, whereas our method requires a larger number of principal directions to explain the same proportion of variance. This suggests that our representation is not only less concentrated in the original channel basis, but also less concentrated in the transformed principal subspace.

Taken together, these results indicate that the learned representation of our method is more distributed both at the channel level and at the subspace level. Importantly, this observation should not be interpreted as stronger cross-view alignment. In fact, our region-level similarity results show that the intermediate cross-view feature similarity of our method is lower than that of the baseline. Therefore, the advantage of our method is better understood not as learning a more compact shared representation, but as preserving a broader and less compressed feature space, which is more compatible with retaining complementary information for downstream collaborative fusion.

\subsection{Broader Impact and Safety Considerations}
\label{app:impact}

The proposed framework aims to improve perception reliability in vehicle-infrastructure collaborative systems, which may contribute to enhanced safety and traffic efficiency in autonomous driving.

Potential risks mainly arise from the deployment of V2X systems in real-world environments. In particular, cross-agent communication and data sharing may introduce privacy and security concerns. In addition, communication instability or synchronization errors may affect the reliability of collaborative perception and influence downstream decision-making modules.

To mitigate these risks, future work should focus on improving system reliability, incorporating uncertainty estimation, and enforcing appropriate data governance and privacy protection mechanisms.

\subsection{Limitations and Future Work}
\label{app:limitations}

Despite the strong performance, the proposed method has several limitations.

\noindent\textbf{Time synchronization.}
In this paper, we do not explicitly discuss the effects of communication latency and temporal misalignment among different agents. In practical V2X systems, observations from different agents may not be perfectly synchronized due to transmission delays and clock offsets. Such asynchrony can introduce feature inconsistency and potentially degrade fusion performance. Therefore, handling asynchronous inputs and developing robust temporal alignment mechanisms remain important directions for future work.

In our framework, the ego-centric fusion architecture may provide partial robustness to temporal misalignment. First, our feature- and attention-based fusion mechanism can adaptively aggregate information from different agents, making the model less sensitive to noisy or inconsistent features caused by temporal offsets. Second, the impact of data asynchrony could be further mitigated by incorporating feature-flow prediction, delay-aware positional encoding, or temporal-aware attention mechanisms. These extensions would allow the model to better compensate for motion-induced feature displacement and explicitly reason about temporal offsets across agents.

\noindent\textbf{Limited modality coverage.}
Our method is designed and evaluated primarily on LiDAR-based inputs. Although DINOv2-based semantic representations are incorporated as auxiliary BEV features, the framework does not explicitly exploit raw RGB observations or perform image-level multi-modal fusion. Extending the framework to incorporate richer multi-modal cues such as joint camera-LiDAR modeling is a promising direction for future research.

%%%%%%%%%%%%%%%%%%%%%%%%%%%%%%%%%%%%%%%%%%%%%%%%%%%%%%%%%%%%

\newpage
\section*{NeurIPS Paper Checklist}

\begin{enumerate}

\item {\bf Claims}
    \item[] Question: Do the main claims made in the abstract and introduction accurately reflect the paper's contributions and scope?
    \item[] Answer: \answerYes{} % Replace by \answerYes{}, \answerNo{}, or \answerNA{}.
    \item[] Justification: %\justificationTODO{}
    The claims in the abstract are consistent with the contributions of the paper. The proposed framework, including DINOv2-based feature extraction, multi-scale BEV fusion, and ego-centric collaboration, is clearly described and implemented. The performance claims, including improved detection accuracy and collaborative gains, are supported by experimental results on a public benchmark, without overgeneralization beyond the evaluated settings.

\item {\bf Limitations}
    \item[] Question: Does the paper discuss the limitations of the work performed by the authors?
    \item[] Answer: \answerYes{} % Replace by \answerYes{}, \answerNo{}, or \answerNA{}.
    \item[] Justification: %\justificationTODO{}
    We discuss the limitations in Sec.~\ref{app:limitations}.

\item {\bf Theory assumptions and proofs}
    \item[] Question: For each theoretical result, does the paper provide the full set of assumptions and a complete (and correct) proof?
    \item[] Answer: \answerNA{} % Replace by \answerYes{}, \answerNo{}, or \answerNA{}.
    \item[] Justification: %\justificationTODO{}
    This paper is primarily an empirical systems and model design study for collaborative 3D object detection. It does not present formal theoretical results, theorems, or mathematical proofs requiring assumptions or derivations.

    \item {\bf Experimental result reproducibility}
    \item[] Question: Does the paper fully disclose all the information needed to reproduce the main experimental results of the paper to the extent that it affects the main claims and/or conclusions of the paper (regardless of whether the code and data are provided or not)?
    \item[] Answer: \answerYes{} % Replace by \answerYes{}, \answerNo{}, or \answerNA{}.
    \item[] Justification: %\justificationTODO{}
    The paper provides detailed descriptions of the model architecture, feature fusion modules, training setup, voxelization parameters, DINOv2 configuration, evaluation protocols, and dataset settings in Sec.~\ref{sec:exp_setup} and App.~\ref{sec:dataset}. These details are sufficient to reproduce the reported main experimental results.

\item {\bf Open access to data and code}
    \item[] Question: Does the paper provide open access to the data and code, with sufficient instructions to faithfully reproduce the main experimental results, as described in supplemental material?
    \item[] Answer: \answerNo{} % Replace by Yes\answerYes{}, \answerNo{}, or \answerNA{}.
    \item[] Justification: %\justificationTODO{}
    The datasets used in this work are publicly available. We provide anonymized supplementary materials with partial code to support reproducibility. However, the complete codebase is not publicly released at submission time and will be made publicly available upon acceptance.
    % The datasets used in this work are publicly available. The supplementary material includes anonymized main experimental result files and an early partial implementation of the proposed method. However, the submitted supplementary material does not constitute a fully self-contained reproduction pipeline, and the complete codebase is not released at submission time. The complete implementation, including code and models, will be made publicly available upon acceptance.

\item {\bf Experimental setting/details}
    \item[] Question: Does the paper specify all the training and test details (e.g., data splits, hyperparameters, how they were chosen, type of optimizer) necessary to understand the results?
    \item[] Answer: \answerYes{} % Replace by \answerYes{}, \answerNo{}, or \answerNA{}.
    \item[] Justification: %\justificationTODO{}
    The paper specifies the experimental setup, including point cloud range, voxel size, BEV projection resolution, pretrained backbone configuration, evaluation metrics, and dataset splits in Sec.~\ref{sec:exp_setup} and App.~\ref{sec:dataset}, and we will make our code publicly available upon acceptance.

\item {\bf Experiment statistical significance}
    \item[] Question: Does the paper report error bars suitably and correctly defined or other appropriate information about the statistical significance of the experiments?
    \item[] Answer: \answerNo{} % Replace by \answerYes{}, \answerNo{}, or \answerNA{}.
    \item[] Justification: %\justificationTODO{}
    The reported results are based on standard benchmark evaluation metrics, but the paper does not report error bars or repeated-run statistics since it is time-consuming to conduct experiments of 3D object detection on large scale datasets.

\item {\bf Experiments compute resources}
    \item[] Question: For each experiment, does the paper provide sufficient information on the computer resources (type of compute workers, memory, time of execution) needed to reproduce the experiments?
    \item[] Answer: \answerYes{} % Replace by \answerYes{}, \answerNo{}, or \answerNA{}.
    \item[] Justification: %\justificationTODO{}
    We provide information about the experimental compute resources in App.~\ref{sec:compute_resources}.
    
\item {\bf Code of ethics}
    \item[] Question: Does the research conducted in the paper conform, in every respect, with the NeurIPS Code of Ethics \url{https://neurips.cc/public/EthicsGuidelines}?
    \item[] Answer: \answerYes{} % Replace by \answerYes{}, \answerNo{}, or \answerNA{}.
    \item[] Justification: %\justificationTODO{}
    The research uses publicly available datasets and standard experimental protocols for collaborative perception. To the best of our knowledge, all aspects of the work conform to the NeurIPS Code of Ethics.

\item {\bf Broader impacts}
    \item[] Question: Does the paper discuss both potential positive societal impacts and negative societal impacts of the work performed?
    \item[] Answer: \answerYes{} % Replace by \answerYes{}, \answerNo{}, or \answerNA{}.
    \item[] Justification: 
    We discuss both potential positive societal impacts and negative societal impacts in Sec.~\ref{app:impact}.
    
\item {\bf Safeguards}
    \item[] Question: Does the paper describe safeguards that have been put in place for responsible release of data or models that have a high risk for misuse (e.g., pre-trained language models, image generators, or scraped datasets)?
    \item[] Answer: \answerNA{} % Replace by \answerYes{}, \answerNo{}, or \answerNA{}.
    \item[] Justification: %\justificationTODO{}
    Our work does not release high-risk generative models, scraped datasets, or other assets that pose significant misuse risks requiring specific safeguards.

\item {\bf Licenses for existing assets}
    \item[] Question: Are the creators or original owners of assets (e.g., code, data, models), used in the paper, properly credited and are the license and terms of use explicitly mentioned and properly respected?
    \item[] Answer: \answerYes{} % Replace by \answerYes{}, \answerNo{}, or \answerNA{}.
    \item[] Justification: %\justificationTODO{}
    All datasets and pretrained models used in this work are properly cited, and their licenses and terms of use are described in App.~\ref{appendix:license}.

\item {\bf New assets}
    \item[] Question: Are new assets introduced in the paper well documented and is the documentation provided alongside the assets?
    \item[] Answer: \answerNo{} % Replace by \answerYes{}, \answerNo{}, or \answerNA{}.
    \item[] Justification: %\justificationTODO{}
    We provide anonymized supplementary materials to support reproducibility. However, the complete codebase and trained models are not publicly released at submission time, and full documentation for the new assets will be provided upon acceptance.
    % We will release the code and models upon acceptance. The supplementary material includes anonymized main experimental result files and an early partial implementation for reference. However, at the time of submission, the new assets are not yet publicly available, and thus no accompanying documentation is provided.

\item {\bf Crowdsourcing and research with human subjects}
    \item[] Question: For crowdsourcing experiments and research with human subjects, does the paper include the full text of instructions given to participants and screenshots, if applicable, as well as details about compensation (if any)? 
    \item[] Answer: \answerNA{} % Replace by \answerYes{}, \answerNo{}, or \answerNA{}.
    \item[] Justification: %\justificationTODO{}
    Our work does not involve crowdsourcing experiments or research with human subjects.

\item {\bf Institutional review board (IRB) approvals or equivalent for research with human subjects}
    \item[] Question: Does the paper describe potential risks incurred by study participants, whether such risks were disclosed to the subjects, and whether Institutional Review Board (IRB) approvals (or an equivalent approval/review based on the requirements of your country or institution) were obtained?
    \item[] Answer: \answerNA{} % Replace by \answerYes{}, \answerNo{}, or \answerNA{}.
    \item[] Justification: %\justificationTODO{}
    Our work does not involve human subjects research requiring IRB approval or equivalent ethical review.

\item {\bf Declaration of LLM usage}
    \item[] Question: Does the paper describe the usage of LLMs if it is an important, original, or non-standard component of the core methods in this research? Note that if the LLM is used only for writing, editing, or formatting purposes and does \emph{not} impact the core methodology, scientific rigor, or originality of the research, declaration is not required.
    %this research? 
    \item[] Answer: \answerNA{} % Replace by \answerYes{}, \answerNo{}, or \answerNA{}.
    \item[] Justification: %\justificationTODO{}
    Large language models are not used as part of the proposed methodology, experiments, or scientific analysis. Any potential use of LLMs for language editing does not affect the core research content.

\end{enumerate}

\end{document}